# Separated Inter/Intra-Modal Fusion Prompts for Compositional Zero-Shot Learning

Sua Jung

Hanyang University

suajung@hanyang.ac.kr

## ABSTRACT

*Compositional Zero-Shot Learning (CZSL) aims to recognize subtle differences in meaning or the combination of states and objects through the use of known and unknown concepts during training. Existing methods either focused on prompt configuration or on using prompts to tune the pre-trained Vision-Language model. However, these methods faced challenges in accurately identifying subtle differences in meaning or combining states with objects. To jointly eradicate the above issues and construct an efficient and effective CZSL technique, we suggest a method to improve attribute recognition performance by utilizing diverse Prompt Learning with an Inter/Intra-Modality Fusion Synthesizer in scene understanding involving subtle semantic differences and multiple objects.*

## KEYWORDS

*Attribute recognition, Prompt-based Learning, Zero-shot Learning*

## 1. INTRODUCTION

When encountering a new thing, such as a blue cat, people often attempt to name it despite the challenge of linking "blue" and "cat'" together. Compositional Zero-Shot Learning (CZSL) aims to recognize and distinguish new concepts. In addition, a detailed textual analysis is needed to understand the semantic differences. Then, it leverages visual information for matching. Previous researches [19], [10], [14], and [26] have focused on connecting attributes with object information to improve prediction accuracy for the final pair. After emerging CLIP [2], researchers use a compatible encoder and prompt, leading to better performance. [16], [22], [9] use soft prompts for pair labels, compared to hard prompts.

A recent paper, DFSP [7], applies a pairing prompt that combined a soft and hard prompt composed of learnable tokens and ``a photo of [state] [object]." However, it struggles to capture subtle language differences using fixed prompts for pairs. Therefore, we develop a method for a better understanding of the connection and semantic differences of vision and language using hard and soft prompts. We conjecture those techniques are too simple to capture all the complexities in visuals and text, and aimed to develop a better method to understand the connection between vision and language for CZSL. We propose a learning method that applies hard and soft prompts to understand semantic differences.

Instead of using just pair prompts, we suggest separating them into attribute, object, and pair prompts. In addition, we compared all possible prompt forms, i.e., hard, hard & soft, and soft prompt in Table III. The features are updated through the suggested Modal Fusion Synthesizer Block (MFSB) to better understand complex intrinsic relationships. Decomposed text and image features based on separated prompts can be helpful rich information to understand complex scenes. We introduce Inter & Intra-modal fusion and Separated Prompts to improve CZSL with prompt learning techniques.

Compared to DFSP [7], hard prompts are only applied for pairs, and the decomposed state and object prompts are in a soft form. It failed to grasp the subtle differences in Figure 3. There was a need for a prompt that focused on states and objects not covered by the current pair prompt. The proposed method uses visual representations for various conditions, with separate state and object settings. Our model outperforms traditional methods, with a significant improvement over baselines. The main contributions are: 1) A suggested framework called Modal Fusion Synthesize Block (MFSB) uses the most optimal detached prompts: Pair, Object, and Attribute. 2) Compared various Modality Fusion and refined features using Separated prompts and Cross-Attention. 3) Several experiments demonstrate that MFSB with Separated Prompts outperforms CZSL benchmarks.

## 2. RELATED WORKS

### 2.1. Compositional Zero-Shot Learning

Compositional Zero-shot Learning (CZSL) improves traditional zero-shot learning by identifying new unseen classes during training. The model can learn new classes by generalizing to unseen attribute combinations with algorithm [1], ALE. CZSL is necessary because new classes keep appearing, making it impractical to retrain models for each one. OADis [20] and others [29], [14], [21] suggested decomposing the encoder and decoder for each state, object, and pair with seen and unseen feature. DFSP [8] others [12], [9] use fusion prompt with visual and linguistic content before predicting. The paper demonstrated the importance of utilizing different types of prompts, such as pair, object, and state, to identify the optimal form, as well as enhancing performance through modal fusion techniques

### 2.2. Prompt-based Learning

The prompt has been utilized as a tool for fine-tuning in NLP [24]. Multiple prompt-based learning researches are done, for example, Visual Prompt [6] suggests the way to use the prompt in the computer-vision field. Red circle [22] has shown a remarkable visual prompt in fine-tuning with only the insertion of the red circle. Those modify prompts for the model instead of adjusting the entire architecture, which is more efficient. In CLIP [3] fine-tuning tasks, previous works adjusted the prompts to guide the model toward image classification goals. Prompt tuning can quickly adjust pre-trained models to diverse tasks without extra training. However, the research concerning the identification of an optimal prompt for CZSL is currently lacking, indicating the importance of this study.

## 3. PROPOSED METHOD

### 3.1. Preliminaries

Compositional Zero-Shot Learning (CZSL) is the task of recognizing seen and unseen compositions for state and object. The baseline is DFSP [8], which implemented the decomposing and fusing module for each vision and text feature. To obtain a text feature, DFSP only uses a hard prompt that contains both state and object information as an input of the CLIP text encoder. DFSP [8] simply fused image and text features using a prompt to narrow the gap for unseen. However, it was not enough to understand the subtle differences in visual and textual levels with only a simple pair prompt. [8] The use of single fusion methods in conjunction with paired data may present limitations in bridging the modality gap when attempting to comprehend language without separation.

Thus, ours is divided in order to enhance clarity and understanding. Also, hard prompts and soft prompts possess distinct perspectives and characteristics, separate prompts are utilized to extract various types of linguistic features in order to enhance the understanding of these prompts, as shown in Table III. Our methodology is inspired by Human-Object Interaction techniques [26], [7], which compose a prompt using a bounding box and category features as a prompt, leading to a better performance. Therefore, we improve the feature diversity that covers subtle differences for unseen domains by using multiple prompts that cover different types of information, leading to a better understanding of subtle joint representation.

As a primitive setting, we define a state set ($A = \{s_0, s_1, ..., s_n\}$) and an object set ($O = \{o_0, o_1, ..., o_n\}$). Based on these two sets, a composition set $C = A \times O$ with size $n \times m$ and subsets of the composition set $C, C^s \cap C^o = \emptyset$. The training set is represented as $T = \{(x_i, c_i) | x \subset X, c \subset C^s\}$, with input image $X$ and seen composition label set $C^s$.

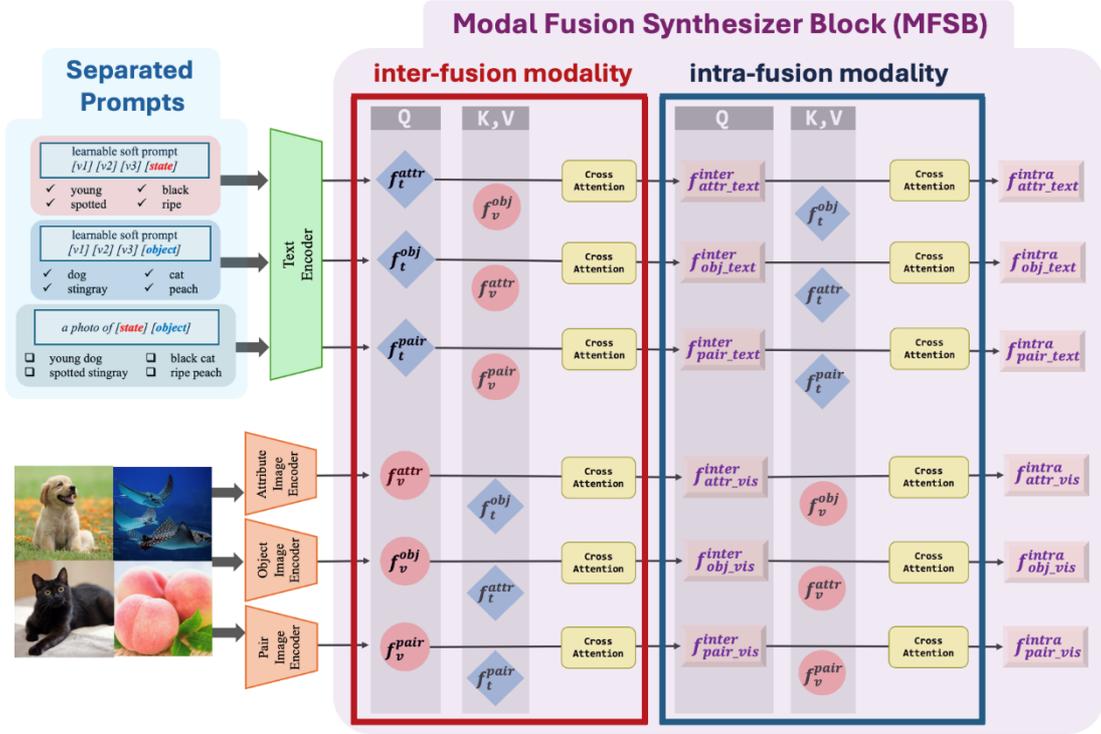

Figure 1. Overall Structure

The description of MFSB includes both Inter-modality fusion, Intra-modality fusion, and distinguished text prompts for each element. Visual features are represented in circular form, while textual features are depicted in diamond form. Using the Inter-modality fusion modality, the textual content can strengthen awareness of each visual feature, object, and state. Through the Intra-modality fusion modality, the output of the Inter-modality fusion modality can exchange the same type of feature content but has common points. In visual, a different object but the same state is inserted as key and value, and in textual, the object that is most relevant to state labels is inserted as key and value. The order of process is Inter first, and then Intra-modality fusion.

### 3.2. Hard Pair Prompt

As used in DFSP [8], hard pair prompt is denoted as: $P_{pair}^{hard} = \{a\ photo\ of, x_s, x_o\}$.

The prompt set is written as $P_{pair}^{hard}$, which is composed of state and object vocabulary, $x_t^s$, $x_t^o$. v and t are image and text vector, which passed $E_v$ image encoder, $E_t$ text encoder, and $\|\cdot\|$ norm calculation. Then, we compute the similarity score for the pair classes: $p_{pair}^{hard} = sim(v_{pair}^{hard}, t_{pair}^{hard})$ where $sim(\cdot,\cdot)$ is a cosine similarity between vision and language vectors.

After matching vision-language followed by the softmax function to obtain p, the cross-entropy loss is used to train the model: $L_{pair}^{hard} = CE(p_{pair}^{hard}, y = (s, o))$

### 3.3. Soft Separated Prompts

Unlike previous CZSL [20], [27] studies, soft prompts have been applied to understand specific content to diverse features in CZSL [18], [8]. CoOp [30] has also improved recognition results by replacing hard with soft prompts. Additionally, we hypothesize that the pair prompt alone is not enough to extract information from a complex scene. Upon presenting the soft separated prompt, it has been confirmed that soft yielded better outcomes, while hard prompt has shown the best performance for a pair in Table III. In ours, the decomposition strategy for only state and object prompt differentiates information about individual elements, exchanging each information for detailed understanding, referring to Table IV The soft separated prompt combines visual and language data for objects and visual and textual features for attributes.

#### 3.3.1. Soft Attribute Prompt

It is composed of learnable vectors and labels, and the prompt set is denoted as: $P_{attr}^{soft} = \{x_0, x_1, ..., x_p, x_s\}$, and $x_0, ..., x_p$ is a prefix content, and $x_s$ is a state vocabulary. From the above, the matching scores are: $p_{attr}^{soft} = sim(v_{attr}^{soft}, t_{attr}^{soft})$. Based on extracted language and image features being matched, the class probability of Separated Attribute Prompt is p and cross-entropy loss is measured with the target attribute class and denoted as: $L_{attr}^{soft} = CE(p_{attr}^{soft}, y = (s))$.

#### 3.3.2. Soft Object Prompt

It is composed of learnable vectors and labels, and the prompt set is denoted as: $P_{obj}^{soft} = \{x_0, x_1, ..., x_p, x_o\}$, and $x_0, ..., x_p$ is a prefix content, and $x_o$ is an object vocabulary. Then the matching score is: $p_{obj}^{soft} = sim(v_{obj}^{soft}, t_{obj}^{soft})$. From the above, the cross-entropy loss is measured with the target object class, and denoted as: $L_{obj}^{soft} = CE(p_{obj}^{soft}, y = (o))$.

### 3.4. Modal Fusion Synthesizer Block (MFSB)

MFSB improves state and object features by combining and breaking down text and images through two types of fusion. First, combine vision and text features through fusion: Inter-modality fusion. Secondly, same-type fusion, Intra-modality fusion. The best choice for refining a feature is to first focus on Table V, Inter- and then Intra-modality fusion. It guides object, attribute, and pairing information in fusion and decomposition stages and improves understanding of information processing over time. In each fusion expression, $a$ and $b$ alternate between vision and text, and $v$ and $t$ are images and text vectors, respectively.

#### 3.4.1. Inter-modality fusion

In the Inter-modality fusion, cross-attention mechanisms combine decomposed text and visual features to merge previously separate text and image features. The visuals and text are enhanced to complement each other, and denoted as:

$$v_{pair}^{inter} = CA(v_{pair}^{hard}, t_{pair}^{hard}, t_{pair}^{hard}), \; t_{pair}^{inter} = CA(t_{pair}^{hard}, v_{pair}^{hard}, v_{pair}^{hard}),$$

$$v_a^{inter} = CA(v_a^{soft}, t_b^{soft}, t_b^{soft}), \; t_a^{inter} = CA(t_a^{soft}, v_b^{soft}, v_b^{soft}).$$

It includes a process where text and visual features are matched for cross-attention. If the query feature is visual, the key and value must be textual. Then, the output is used as an input for the Inter-modality fusion modality.

### 3.4.1.1. Refined Pair Prompt Features

Since hard pair prompt is being used, so the hard pair textual feature $t_{pair}^{hard}$ incorporates with visual feature $v_{pair}^{hard}$ through MFSB's Inter-modality fusion. After fusion and decomposition of the $\{state, object\}$ set, a refined pair prompt feature with updated information is created. Then, the matching score with the ground-truth pair is $p_{pair}^{inter} = sim(v_{pair}^{hard}, t_{pair}^{hard})$, and calculated cross-entropy inter-pair loss is: $L_{pair}^{inter} = CE\left(p_{pair}^{inter}, \; y = (s, o)\right)$.

### 3.4.1.2. Refined Attribute Prompt Features

Similar with above, soft attribute prompt is being used, so the soft attribute textual feature $t_{attr}^{soft}$ incorporates with visual feature $v_{attr}^{soft}$ through MFSB's Inter-modality fusion. After fusion and decomposition, a refined attribute prompt feature is created. The matching score with the ground-truth attribute is: $p_{attr}^{inter} = sim(v_{attr}^{soft}, t_{attr}^{soft})$. Then, we calculate cross-entropy inter-attribute loss as: $L_{attr}^{inter} = CE\left(p_{attr}^{inter}, \; y = (s)\right)$.

### 3.4.1.3. Refined Object Prompt Features

As mentioned, soft object prompt is being used, so the soft object textual feature $t_{attr}^{soft}$ incorporates with visual feature $v_{obj}^{soft}$ through MFSB's Inter-modality fusion. After fusion and decomposition, a refined object prompt feature is created. From the above, the matching score with the ground-truth object is $p_{obj}^{inter} = sim(v_{obj}^{soft}, t_{obj}^{soft})$. We then calculate cross-entropy inter-object loss as: $L_{obj}^{inter} = CE\left(p_{obj}^{inter}, \; y = (o)\right)$.

### 3.4.2. Intra-modality fusion

For the Intra-modality fusion modality, the output of Inter-modality fusion modality features is being used as an input. And those are denoted as:

$$v_{pair}^{intra} = CA(v_{pair}^{inter}, t_{pair}^{inter}, t_{pair}^{inter}), \; t_{pair}^{intra} = CA(t_{pair}^{inter}, v_{pair}^{inter}, v_{pair}^{inter}),$$

$$v_a^{intra} = CA(v_a^{inter}, t_b^{inter}, t_b^{inter}), \; t_a^{intra} = CA(t_a^{inter}, v_b^{inter}, v_b^{inter}).$$

In this case, the inter-fused feature is used for cross-attention with matching key and value features of the same type. For example, when a visual feature of an object is queried, the key and value are attributes of the visual feature.

### 3.4.2.1. Intra-modality fusion #1. ATTR ← OBJ

Intra-modality fusion #1) applies a cross-attention [5], CA. As a query, inter-fused soft attribute prompt visual feature $v_{attr}^{inter}$ is inserted and inter object prompt feature $v_{obj}^{inter}$ is used as key and value. In text-level, inter attribute prompt textual feature $t_{attr}^{inter}$ and inter soft object textual feature $t_{obj}^{inter}$ are applied. From these, attribute information can be redeemed by using related object content as a reference. Those are described as:

$$v_{attr}^{intra} = CA(v_{attr}^{inter}, t_{obj}^{inter}, t_{obj}^{inter}), \quad t_{attr}^{intra} = CA(t_{attr}^{inter}, v_{obj}^{inter}, v_{obj}^{inter}).$$

The matching score between vision and language is $p_{attr}^{intra} = sim(v_{attr}^{intra}, t_{attr}^{intra})$.. We calculate cross-entropy intra-attribute loss is: $L_{attr}^{intra} = CE\left(p_{attr}^{intra}, y = (s)\right)$.

### 3.4.2.2. Intra-modality fusion #2. OBJ ← ATTR

Intra-modality fusion #2) also utilizes CA, but vice-versa. As a query, inter-fused soft object prompt visual feature $v_{obj}^{inter}$ is inserted and inter attribute prompt feature $v_{attr}^{inter}$ is used as key and value. In text-level, inter object prompt textual feature $t_{obj}^{inter}$ and inter soft attribute textual feature $t_{attr}^{inter}$ are applied. Those are described as:

$$v_{obj}^{intra} = CA(v_{obj}^{inter}, t_{attr}^{inter}, t_{attr}^{inter}), \quad t_{obj}^{intra} = CA(t_{obj}^{inter}, v_{attr}^{inter}, v_{attr}^{inter}).$$

The matching score between vision-language is $p_{obj}^{intra} = sim(v_{obj}^{intra}, t_{obj}^{intra})$, and calculated cross-entropy intra-object loss is: $L_{obj}^{intra} = CE\left(p_{obj}^{intra}, y = (o)\right)$.

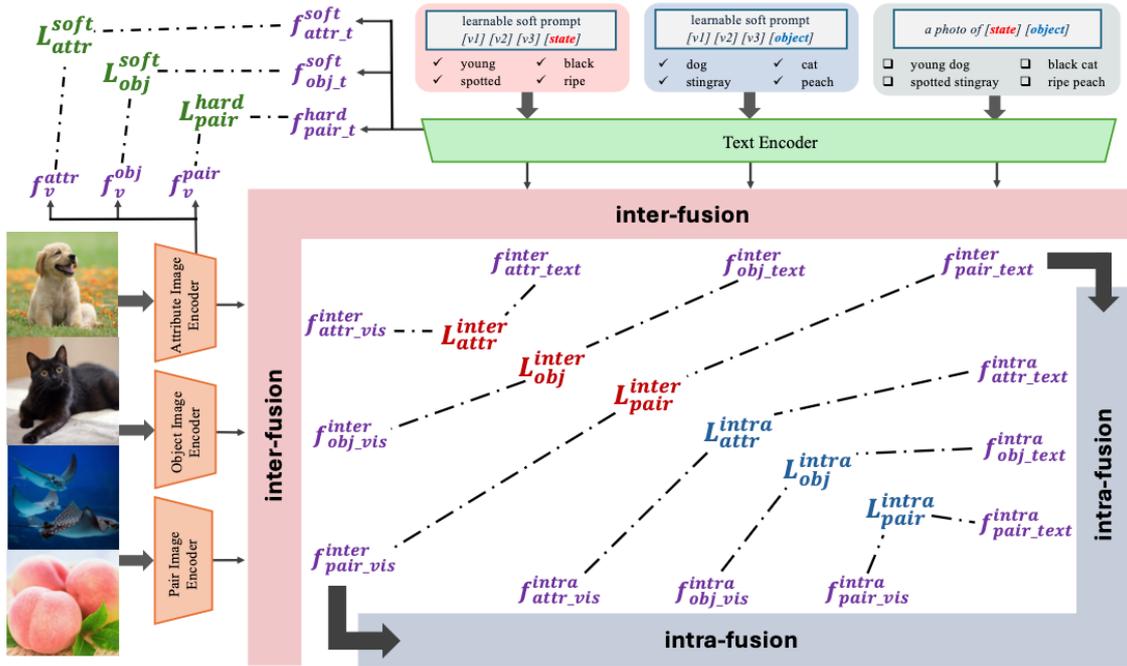

Figure 2. Matching Structure for loss in each feature element

### 3.5. Total Loss

To understand how visual and textual prompts relate, the total loss is calculated by combining multiple losses. The types of loss in this paper are 1) Hard Pair, 2) Soft Attr & Obj, 3) Inter-fused Pair & Attr & Obj, 4) Intra-fused Pair & Attr & Obj. $L_{pair}^{hard+soft}$ indicates an existing loss from DFSP, baseline. $L_{attr}^{hard+soft}$ and $L_{obj}^{hard+soft}$ are added additionally based on $L_{pair}^{hard+soft}$ format.

The total loss is:

$$L_{total} = L_{pair}^{hard+soft} \times 0.1 + \left(L_{attr}^{hard+soft} + L_{obj}^{hard+soft}\right) \times 0.01$$
$$+ \left(L_{pair}^{hard} + L_{attr}^{soft} + L_{obj}^{soft}\right) \times \alpha$$
$$+ \left(L_{pair}^{inter} + L_{attr}^{inter} + L_{obj}^{inter}\right) \times \beta$$
$$+ \left(L_{pair}^{intra} + L_{attr}^{intra} + L_{obj}^{intra}\right) \times \gamma,$$

where α, β, and γ are hyper-parameters set as 0.2.

## 4. EXPERIMENTS

### 4.1. Dataset and Details

We used three challenging benchmark datasets: MIT-States [4], UT-Zappos [28], and C-GQA [13] in Closed and Open world settings. MIT-States has 53,753 images with 245 objects and 115 attributes. MIT-States [4] has 300 unseen and 1262 seen labels for validation, and 400 unseen labels for the test set. The dataset has 50,025 shoe images labeled under 12 categories and 16 attributes. In a closed-world scenario, there are 15 unseen validation instances, 18 unseen test instances, and 83 seen labels. The C-GQA [13] dataset has 39,298 images with 870 object labels and 453 attribute labels. Metrics assess accuracy for both seen and unseen compositions. Accuracy is measured on seen and unseen categories in both Seen (S) and Unseen (U) data. The Harmonic Mean (HM) metric estimates total accuracy. The AUC metric measures the balance of true positive and false positive rates at varying decision thresholds. Our method is implemented with PyTorch [19] 1.12.1 and Adam [2] optimizer is utilized for optimization. Each three challenging datasets is trained for 20 epochs, and both image and text encoder are based on pre-trained CLIP [3] Vit-L/14 model with 1 × NVIDIA RTX 4090 GPU.

Table 1. Open-World Setting Results

| Method | MIT-States | | | | UT-Zappos | | | | CGQA | | | |
|---|---|---|---|---|---|---|---|---|---|---|---|---|
| | S | U | HM | AUC | S | U | HM | AUC | S | U | HM | AUC |
| AoP[16] | 16.6 | 5.7 | 4.7 | 0.7 | 50.9 | 34.2 | 29.4 | 13.7 | - | - | - | - |
| LE+[12] | 14.2 | 2.5 | 2.7 | 0.3 | 60.4 | 36.5 | 30.5 | 16.3 | 19.2 | 0.7 | 1.0 | 0.08 |
| TMN[21] | 12.6 | 0.9 | 1.2 | 0.1 | 55.9 | 18.1 | 21.7 | 8.4 | - | - | - | - |
| SymNet[27] | 21.4 | 7.0 | 5.8 | 0.8 | 53.3 | 44.6 | 34.5 | 18.5 | 26.7 | 2.2 | 3.3 | 0.43 |
| CompCos[9] | 25.4 | 10.0 | 8.9 | 1.6 | 59.3 | 46.8 | 36.9 | 21.3 | - | - | - | - |
| CGE[11] | 32.4 | 5.1 | 6.0 | 1.0 | 61.7 | 47.7 | 39.0 | 23.1 | 32.1 | 2.0 | 3.4 | 0.5 |
| Co-CGE[10] | 31.1 | 5.8 | 6.4 | 1.1 | 62.0 | 44.3 | 40.3 | 23.1 | 32.1 | 2.0 | 3.4 | 0.5 |
| KG-SP[23] | 28.4 | 7.5 | 7.4 | 1.3 | 61.8 | 52.1 | 42.3 | 26.5 | 31.5 | 2.9 | 4.7 | 0.78 |
| CSP[17] | 46.3 | 15.7 | 17.4 | 5.7 | 64.1 | 44.1 | 38.9 | 22.7 | 28.7 | 5.2 | 6.9 | 1.20 |
| DFSP[8] | 47.4 | 18.1 | 19.1 | 6.7 | 63.5 | 53.8 | 41.2 | 26.4 | 35.6 | 6.5 | 9.0 | 1.95 |
| **Ours** | **49.33** | **19.01** | **20.35** | **7.33** | **65.24** | **55.0** | **43.21** | **29.27** | **39.20** | **7.25** | **11.26** | **2.90** |

Quantitative results on three challenging benchmark datasets in Open-World Setting. Our final model with Seen(S), Unseen(U), AUC, and HM shows the best performance in most of the classes.

Table 2. Closed-World Setting Results

| Method | MIT-States | | | | UT-Zappos | | | | CGQA | | | |
|---|---|---|---|---|---|---|---|---|---|---|---|---|
| | S | U | HM | AUC | S | U | HM | AUC | S | U | HM | AUC |
| AoP[16] | 14.3 | 17.4 | 9.9 | 1.6 | 59.8 | 54.2 | 40.8 | 25.9 | 17.0 | 5.6 | 5.9 | 0.7 |
| LE+[12] | 15.0 | 20.1 | 10.7 | 2.0 | 53.0 | 61.9 | 41.0 | 25.7 | 18.1 | 5.6 | 6.1 | 0.8 |
| TMN[21] | 20.2 | 20.1 | 13.0 | 2.9 | 58.7 | 60.0 | 45.0 | 29.3 | 23.1 | 6.5 | 7.5 | 1.1 |
| SymNet[27] | 24.2 | 25.2 | 16.1 | 3.0 | 49.8 | 57.4 | 40.4 | 23.4 | 26.8 | 10.3 | 11.0 | 2.1 |
| CompCos[9] | 25.3 | 24.6 | 16.4 | 4.5 | 59.8 | 62.5 | 43.1 | 28.1 | 28.1 | 11.2 | 12.4 | 2.6 |
| CGE[11] | 28.7 | 25.3 | 17.2 | 5.1 | 56.8 | 63.6 | 41.2 | 26.4 | 28.7 | 25.3 | 17.2 | 5.1 |
| Co-CGE[10] | 31.1 | 5.8 | 6.4 | 1.1 | 62.0 | 44.3 | 40.3 | 23.1 | 32.1 | 2.0 | 3.4 | 0.5 |
| SCEN[25] | 29.9 | 25.2 | 18.4 | 5.3 | 63.5 | 63.1 | 47.8 | 32.0 | 28.9 | 25.4 | 17.5 | 5.5 |
| CSP[17] | 46.6 | 49.9 | 36.3 | 19.4 | 64.2 | 66.2 | 46.6 | 33.0 | 28.8 | 26.8 | 20.5 | 6.2 |
| DFSP[8] | 46.9 | 52.0 | 37.2 | 20.6 | 63.3 | 66.4 | 45.1 | 32.1 | 35.6 | 29.3 | 24.3 | 8.7 |
| **Ours** | **48.99** | **52.31** | **38.4** | **21.59** | **65.28** | **67.03** | **47.6** | **35.01** | **37.85** | **30.97** | **25.6** | **10.1** |

Quantitative results on three challenging benchmark datasets in Closed-World Setting. Our final model with Seen(S), Unseen(U), AUC, and HM shows the best performance in most of the classes.

### 4.2. Results

We compare our method DFSP [8] with previous compositional zero-shot learning techniques, including AoP [16] and LE+ [12] TMN [21], SymNet [27], Comp-Cos [9], CGE [11], Co-CGE [10], SCEN [25], KG-SP [23], and CSP [17]. We investigate various fusion methods including Separated Prompts and Modal Fusion Synthesizer Block (MFSB), including inter- and intra-fusion modality. Table I and Table II show that our method performs better on MIT-States [4], UT-Zappos [28], and CGQA [13] datasets. Ours outperforms in AUC scores of 5.4% on MIT-States [4], 36.0% on UT-Zappos [28], and 10.5% on CGQA [13], surpassing by 4.3%. Also, there is a 6.8% increase in the harmonic mean on the MIT-States [4] dataset compared to other methods. Also, it shows high accuracy on both seen and unseen datasets in these experiments.

Based on the results, the inter- and intra-fusion with Separated prompts are the most effective for prompt composition in both Open and Closed world settings. Improving image and text features through different fusion types and focusing on attribute or object recognition can yield better results than a basic fusion approach in the language branch. The results show that our method improves model performance for compositional zero-shot learning on three challenging datasets.

### 4.3. Ablation Study

Ablation studies aim to demonstrate the effectiveness of proposed methods and identify the best prompt for CZSL. Table 3 identifies the best prompt combination, and Table 4 compares the components of the prompt. Table 5 demonstrates the comparison of fusion techniques. Every experiment is conducted with the MIT-States dataset in the Open-world setting.

Table 3. Ablation study #1

| Methods | S | U | HM | AUC |
|---|---|---|---|---|
| **Hard+Soft {Pair}, Hard {Obj}, Hard {Attr}** | 48.76 | 18.23 | 19.21 | 6.60 |
| **Hard+Soft {Pair}, Hard {Obj}, Soft {Attr}** | 49.20 | 18.37 | 19.90 | 7.17 |
| **Hard+Soft {Pair}, Hard {Obj}, Hard+Soft {Attr}** | 42.48 | 16.94 | 17.64 | 5.57 |
| **Hard+Soft {Pair}, Soft {Obj}, Hard {Attr}** | 48.91 | 18.31 | 19.48 | 6.85 |
| **Hard+Soft {Pair}, Soft {Obj}, Soft {Attr}** | 48.87 | 18.79 | 20.23 | 7.29 |
| **Hard+Soft {Pair}, Soft {Obj}, Hard+Soft {Attr}** | 42.73 | 17.97 | 18.76 | 6.10 |
| **Hard+Soft {Pair}, Hard+Soft {Obj}, Hard {Attr}** | 46.12 | 17.83 | 18.35 | 6.46 |
| **Hard+Soft {Pair}, Hard+Soft {Obj}, Soft {Attr}** | 48.28 | 18.46 | 20.02 | 7.08 |
| **Hard+Soft {Pair}, Hard+Soft {Obj}, Hard+Soft {Attr}** | 48.03 | 18.49 | 20.08 | 7.04 |
| **Hard {Pair}, Hard {Obj}, Hard {Attr}** | 47.42 | 17.00 | 19.13 | 6.83 |
| **Hard {Pair}, Hard {Obj}, Soft {Attr}** | 48.61 | 18.59 | 20.23 | 7.24 |
| **Hard {Pair}, Hard {Obj}, Hard+Soft {Attr}** | 48.78 | 18.51 | 20.19 | 7.20 |
| **Soft {Pair}, Soft {Obj}, Hard {Attr}** | 47.80 | 17.29 | 19.48 | 6.95 |
| **Soft {Pair}, Soft {Obj}, Soft {Attr}** | 48.66 | 18.53 | 20.28 | 7.28 |
| **Soft {Pair}, Soft {Obj}, Hard+Soft {Attr}** | 46.85 | 18.72 | 19.79 | 6.99 |
| **Soft {Pair}, Hard+Soft {Obj}, Hard {Attr}** | 47.50 | 17.03 | 19.01 | 6.82 |
| **Soft {Pair}, Hard+Soft {Obj}, Soft {Attr}** | 48.11 | 19.01 | 20.30 | 7.28 |
| **Soft {Pair}, Hard+Soft {Obj}, Hard+Soft {Attr}** | 48.03 | 18.84 | 20.23 | 7.19 |
| **Hard {Pair}, Hard {Obj}, Hard {Attr}** | 48.32 | 18.99 | 20.11 | 7.18 |
| **Hard {Pair}, Hard {Obj}, Soft {Attr}** | 49.16 | 18.74 | 20.21 | 7.30 |
| **Hard {Pair}, Hard {Obj}, Hard+Soft {Attr}** | 47.14 | 18.67 | 19.62 | 6.95 |
| **Hard {Pair}, Soft {Obj}, Hard {Attr}** | 48.03 | 18.77 | 20.24 | 7.16 |
| **Hard {Pair}, Soft {Obj}, Soft {Attr}** | **49.33** | **19.01** | **20.35** | **7.33** |
| **Hard {Pair}, Soft {Obj}, Hard+Soft {Attr}** | 46.81 | 17.86 | 18.96 | 6.46 |
| **Hard {Pair}, Hard+Soft {Obj}, Hard {Attr}** | 46.21 | 18.65 | 19.60 | 6.80 |
| **Hard {Pair}, Hard+Soft {Obj}, Soft {Attr}** | 48.91 | 18.55 | 20.31 | 7.27 |
| **Hard {Pair}, Hard+Soft {Obj}, Hard+Soft {Attr}** | 48.32 | 18.74 | 20.05 | 7.16 |

Seen(S), Unseen(U), AUC, and HM results for using prompt setting as Hard+Soft, Hard, and Soft.

**4.3.1. Hard+Soft, Hard, Soft**

An analysis comparing hard and soft prompts is presented in Table III. Previous research often used a mix of hard and soft prompts or just soft or hard prompts for training. However, using hard for a pair and soft for an object/state showed the best performance overall. The hard pair refines information from the inputs to enhance comprehension of the scene. Table III illustrates the importance of appropriately utilizing both hard and soft characteristics based on the label to comprehend complex scenes.

Table 4.  Ablation study #2

| Methods | S | U | HM | AUC |
|---|---|---|---|---|
| Pair | 47.41 | 18.13 | 19.10 | 6.74 |
| Object | 47.59 | 17.99 | 19.37 | 6.69 |
| State | 47.63 | 18.38 | 19.25 | 6.86 |
| Object + State | 47.89 | 18.23 | 19.23 | 6.91 |
| Pair + Object | 47.90 | 18.16 | 19.58 | 7.00 |
| Pair + State | 48.61 | 18.73 | 19.98 | 7.26 |
| **Pair + State + Object** | **49.33** | **19.01** | **20.35** | **7.33** |

Seen(S), Unseen(U), AUC, and HM results for using prompt component as a single one for each and all together.

### 4.3.2. Prompt Component Comparison of three elements

Table IV indicates higher accuracy with three types of prompts compared to a single pair prompt. Prior studies [8], [14], [20] have used the single-pair prompt. Our research found that the best approach uses three separate prompt components, which separate object and state within the pair prompt. The utilization of a specific attribute and an object prompt demonstrated effectiveness in enhancing the comprehension of each element's visual and linguistic content, resulting in increased performance over time. Also, deconstructed prompts helped clarify intricate relationships.

Table 5.  Ablation study #3

| Methods | S | U | HM | AUC |
|---|---|---|---|---|
| No Fusion | 47.41 | 18.13 | 19.10 | 6.74 |
| Intra-Fusion Only | 48.11 | 18.67 | 19.36 | 7.05 |
| Inter-Fusion Only | 48.96 | 18.98 | 19.80 | 7.27 |
| 1. Intra 2. Inter | 48.85 | 18.80 | 20.02 | 7.18 |
| **1. Inter 2. Intra** | **49.33** | **19.01** | **20.35** | **7.33** |

Seen(S), Unseen(U), AUC, and HM results for no fusion, single fusion, and fusion order.

### 4.3.3. Comparison of Inter-and Intra-modality fusion

Table V displays results of no, single, and both but different order of fusions. The results show that MFSB is effective with improved accuracy metrics. Some parameters from separated prompts pass through the MFSB block to access important information for understanding complex scenes. Inter-fused features successfully mix each vision and language content on each perspective, and Intra-fused features updated content that has relevant vice-versa elements. Also, the order after

Intra-modality Fusion for intra-fusion is the most effective to make a response to new unseen compositions in CZSL.

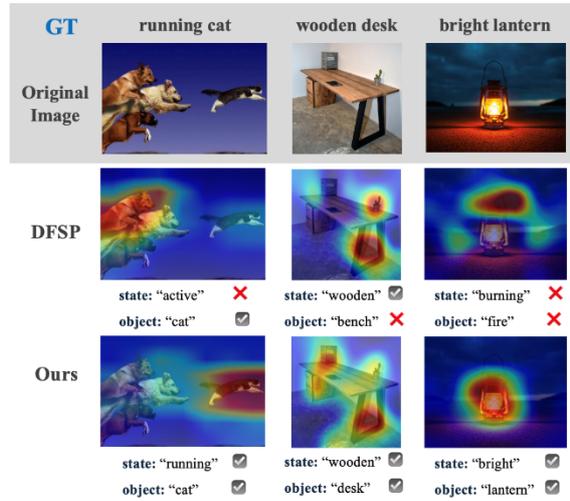

Figure 3. Inference comparison between baseline and ours. As shown, ours can predict accurately by a better understanding in intrinsic relationships.

## 5. CONCLUSION

In this study, we introduce a novel method called Modal Fusion Synthesized Prompt to accurately identify nuanced distinctions in meaning for composing state and object during training. As grounded from prompt-tuning, a hard pair and soft decomposed prompt are applied to diverse usage in training. Furthermore, the proposed prompts serve different purposes in gathering precise and comprehensive information from the training data. The hard pair prompt gathers detailed information about multiple objects in a scene, with separate soft prompts capturing specific features based on complex scenarios and using inter- and intra-modal fusion to improve comprehension of visual and linguistic contexts. Extensive experiments on three challenging datasets demonstrate the effectiveness of our proposed method, Inter- & Intra-Modality Fusion with Separated Prompts.

**Authors**

Sua Jung

: Master Candidate in Hanyang University

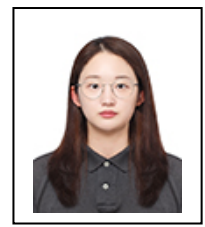